\documentclass{article}
\usepackage{graphicx}
\usepackage{cleveref}
\usepackage{longtable}
\usepackage{pdflscape}
\usepackage{array} 
\usepackage[style=apa]{biblatex}
\usepackage{xcolor}
\title{Word Meanings in Transformer Language Models}
\author{Jumbly Grindrod \thanks{University of Reading, Department of Philosophy} \and Peter Grindrod \thanks{University of Oxford, Mathematical Institute} }
\date{July 2025}
\begin{document}

\maketitle

\begin{abstract}
We investigate how word meanings are represented in the transformer language models. Specifically, we focus on whether transformer models employ something analogous to a lexical store - where each word has an entry that contains semantic information. To do this, we extracted the token embedding space of RoBERTa-base and k-means clustered it into 200 clusters. In our first study, we then manually inspected the resultant clusters to consider whether they are sensitive to semantic information. In our second study, we tested whether the clusters are sensitive to five psycholinguistic measures: valence, concreteness, iconicity, taboo, and age of acquisition. Overall, our findings were very positive - there is a wide variety of semantic information encoded within the token embedding space. This serves to rule out certain "meaning eliminativist" hypotheses about how transformer LLMs process semantic information.

\end{abstract}

\section{Introduction}\label{sec:intro}

Do large language models (LLMs) understand the meanings of the words that they use? When we apply terms like “understand” - terms that are typically applied in the human case - to an artificial system, we inevitably enter into a debate with an anthropomorphised framing. There, the possibility of a skeptical answer looms because LLMs fail to possess some feature that seems important in the human case. An alternative approach is to stipulate that LLMs understand their words in some sense and then  ask what that understanding consists of. We can call it “understanding*” or “AI-understanding” if we like, but for the purposes of this paper we will stick with the original term. Even if attributing this kind of understanding to LLMs does not equate to attributing human understanding, it may be that an investigation into the way LLMs understand the words they use could still prove useful in the human case. One way in which this could occur is that LLM could help answer `how possibly' questions about possible ways in which linguistic information can be processed, even if it is not the way linguistic information is processed in the human case (Grindrod, forthcoming). 

In this paper, we focus specifically on how semantic lexical information is stored and employed within a particular kind of large language model architecture - the transformer architecture \parencite{vaswaniAttentionAllYou2017}. The transformer architecture is one of the reasons for the remarkable progress seen in language model technology and is still the basis for the current state-of-the-art. One of the fascinating aspects about the transformer architecture is that the self-attention mechanism at its heart gives rise to two distinct representations for any given word it processes. On the one hand, there is the “token embedding” or “static embedding” that is invariantly assigned to each word in the LLM’s dictionary and that (once combined with a positional embedding through vector addition) serves as input to the self-attention mechanism. On the other hand, there is the “contextualised embedding” that is the output of the self-attention procedure and that serves as a representation of the word as it was used in the input text. One of the key successes of the transformer architecture is the ability to represent a word as it is used in a particular context, and the contextualised embedding plays this role.

A long-standing debate in philosophy of language concerns the extent to which the meaning of a word as used on a particular occasion is determined by its invariant word meaning on the one hand and the context in which it is used on the other \parencite{wittgensteinPhilosophicalInvestigations1953, travisPragmatics1997, cappelenInsensitiveSemanticsDefense2005, borgMinimalSemantics2004}. The distinction between static and contextualized embeddings within LLMs leads to an analogous question: to what extent is a word’s contextualised embedding determined by the word’s static embedding? One possibility is that the static embeddings are rich with semantic information and that much of this is retained in the contextualized embeddings. Another possibility is that as far as semantic properties go, the static embeddings merely serve as placeholders, with semantic information being introduced somewhere within the self-attention mechanism. 

We approach this question through an empirical investigation of the information stored within the static embeddings. We extract the static embeddings from RoBERTa-base, an open-source model available through Hugging Face’s transformers package \parencite{liuRoBERTaRobustlyOptimized2019}. We perform a cluster analysis on the static embeddings, and then manually inspect the clusters to investigate the static embedding space. We then test for whether the arrangement of the clusters is sensitive to a range of psycholinguistic measures, as a way of testing whether the static embedding space is sensitive to semantic information. Our findings show that the static embedding space is in fact rich with a range of semantic information. LLMs succeed in understanding via the use of a kind of lexical store, where semantic information is encoded for each word in their vocabulary.

The paper is structured as follows. In \cref{sec:Tmawr}, we provide a brief informal overview of the transformer architecture and of the role of static and contextualised embeddings. Then in \cref{sec:Ceac}, we frame our investigation in terms of the radical contextualism debate within philosophy of language. Following \parencite{grindrodTransformersContextualismPolysemyforthcoming}, we show that our investigation will serve to test a position analogous to the “meaning eliminativism”  previously proposed by the likes of \textcite{recanatiLiteralMeaning2003}, \textcite{rayoPleaSemanticLocalism2013}, and \textcite{elmanAlternativeViewMental2004}. We then present our cluster analysis in \cref{sec:study1}, along with the findings of our first study, a manual inspection of the clusters. In \cref{sec:study2}, we present the findings of our second study, where we test whether the clusters are sensitive to a range of psycholinguistic measures.

\section{Transformer models and word representations}\label{sec:Tmawr}

The introduction of the transformer architecture was one of the key developments that led to the progress in artificial intelligence seen in recent years. For our purposes, it will prove useful to understand a little bit about its distinctive features. One key issue in natural language processing has been how to process linguistic data in a way that is sensitive to its wider linguistic context. We could process “cat” in “the cat is on them mat” via use of a single vector that is employed every time the word is used. But only doing that will leave out that the word “cat” plays a particular role in that sentence - for example, it binds with “the” in a particular way to produce a determiner phrase, which then serves as the subject for which “is on the mat” is the predicate. None of this is captured if we associate a single invariant vector with “cat”.

The transformer architecture uses a self-attention mechanism to capture dependencies between datapoints. The self-attention mechanism generates contextualised embeddings on the basis of the static embeddings that are invariantly assigned to each word or sub-word, combined with the manner in which the words reside in their wider linguistic context. This task is split across a number of self-attention heads that through training learn to focus on particular aspects of the wider linguistic context and have this determine the resultant contextualized embedding.\footnote{ For a survey of experimental work investigating the roles of various attention heads, see: \parencite{zhengAttentionHeadsLarge2024}.} This procedure is repeated across many layers, giving the model opportunity to take into account all relevant aspects of the linguistic context in producing a contextualized embedding for each word.\footnote{ RoBERTa has 12 self-attention layers, each with 12 self-attention heads. But this is comparatively small compared to more recent models.}

All aspects of the model, including the static embeddings and the self-attention weights, are generated simultaneously through the same set of training procedures. And while the static embeddings and contextualized embeddings contain relatively few parameters per word (in Roberta-base, the dimensionality of the embeddings is 768), there are many millions of parameters across the entire model, which makes the manner in which the model is able to generate the contextualized embeddings opaque. As such, it is an open empirical question how exactly LLMs are able to perform in the way that they do. Much of the empirical work has focused on behaviour of the self-attention heads and, to a lesser extent, the feed-forward networks that come after each self-attention layer. By contrast, we focus here on the static embeddings that serve as input to the self-attention procedure.

\section{Contextualised embeddings and contextualism}\label{sec:Ceac}

As stated in the introduction, the distinction between the static embedding and the contextualised embedding maps fairly clearly onto an intuitive distinction between a word’s meaning and what a word means when used on a particular occasion. An initially intuitive view is that the relation between these two notions is or approaches identity, that what a word means on a particular occasion of use is just determined completely by its invariant meaning. But this has been challenged most notably by contextualists \parencite{travisPragmatics1997, recanatiLiteralMeaning2003}. They argue that (nearly) all words vary in terms of what they contribute to a sentence meaning on a particular occasion of use, and that there are a wide, possibly open-ended, range of contextual factors that can determine this. 

Assuming that such variation in usage is right, it is then controversial what implications this has for word meaning. Some argue that word meanings are nevertheless rich in information, even if they are subsequently modulated when used. Others argue that word meanings must be some minimal core that is subsequently enriched on each occasion. Perhaps most radically, some have suggested that the notion of a static word meaning is redundant, that utterance meaning can be generated without some dedicated store of semantic information for each word. \textcite{recanatiLiteralMeaning2003} labels such a view “meaning eliminativism”, and it is a view that has arguably been defended in psycholinguistics by \textcite{elmanAlternativeViewMental2004} and in philosophy by \textcite{rayoPleaSemanticLocalism2013}. Within LLMs, there is a straightforward way in which the meaning eliminativist view could be realized \parencite{grindrodTransformersContextualismPolysemyforthcoming}. It may turn out that the static embeddings contain little in the way of semantic information, perhaps they serve as mere placeholders, or perhaps they only contain information about morphology and syntax. But what reason might there be for the LLM to take this approach? On the one hand, we should consider the wide array of information that any given word has associated with it, including morphological, phonological, syntactic, semantic, and pragmatic information. Given this, combined with the fact that the embedding space presumably has a limit on the amount of information it can store regarding each word, it may be that some semantic information, particularly information that is context-sensitive, is really introduced at the self-attention procedure. The relative size of the model speaks in favour of this point as well; as noted earlier, the embeddings for each word form a relatively small parts of the overall model; in RoBERTa-base they have only 768 parameters compared to the 10s of millions of parameters contained within the self-attention and feed-forward layers. 

We can be more specific in our inquiry, however, and ask not only whether there is semantic information contained within the static embeddings but also what kind of semantic information. This is something that we explore through our cluster analysis of RoBERTa-base, which we describe in the rest of the paper.

\section{Study 1: Manual inspection of clusters}\label{sec:study1}

We extracted the static embeddings from RoBERTa-base, a widely-used open source transformer model available through Hugging Face’s ``transformer'' Python library \parencite{liuRoBERTaRobustlyOptimized2019}. The vocabulary size for RoBERTa is 50,265, 
and the dimensionality of the embedding space is 768. We then performed k-means clustering on the space into 200 clusters. The largest cluster is 1,524, 
smallest cluster is 1, and 45 clusters contained 50 or fewer words. 
 We then manually inspected each cluster in order to consider the features that unite the terms within them. 

A significant portion of the clusters are relatively uninteresting. As vocabularies are generated through an automatic tokenization method, much of the vocabulary consists in special symbols and word parts.\footnote{ Liu et al. (2019) employed bite-pair encoding in developing RoBERTa.} As a result, 59 of the clusters exclusively contain such tokens. Many of the smaller clusters (those containing 10 or fewer tokens, of which there are 25) are relatively uninteresting as they are dedicated to specific words or word parts. In total, 74 clusters can be discounted in this way.\footnote{ Although the focus in this study is not on the sub-word parts, there is clearly substantial information that some sub-word part clusters are sensitive to. For instance, cluster 32 contains suffixes for place-names (e.g. “man”, “ville”, “ley”, “ford”) while cluster 57 appears dedicated to high register suffixes (e.g. “ologists”, “ogeneous”, “otechnology”). But for the remainder of the study we will ignore these clusters.}

Beyond these less interesting clusters, manual inspection reveals that there are a number of typological, morphological, and syntactic features that affect the organization of the clusters. On the typographical side, many clusters contain only either capitalized (e.g. cluster 83) or non-capitalized (e.g. cluster 38) terms. On the syntactic side, many clusters contain only terms from a particular POS category (e.g. only adjectives (cluster 3), verbs (cluster 69), adverbs (clusters 78, 127) etc.). On the morphological side, some clusters contain only words that share a common ending e.g. “-ing” (cluster 27) , “-ed” (clusters 55, 121, 128, 189), “-ly” (clusters 78, 127).

More importantly for this study, a large number of clusters appear to be sensitive to the meanings of the terms involved, providing clear evidence that semantic information is encoded within the static embeddings. In \cref{tab:sensitive clusters} we list 67 clusters that are the clearer examples of being united according to the meanings of their words. We have listed them alongside the top 5 terms in the cluster that were closest to the centroid.\footnote{ The intuition behind this ordering is that the words in the centre of the cluster will be more indicative of the theme, but we have not tested this intuition.}$^,$ \footnote{ A “?” prior to the word indicates that the word follows a space. Where words lack a “?”, it indicates either that they are combined with other words or that they appear at the start of a sentence.}

{\small
\begin{landscape} 
\begin{longtable}{lll}                                                                                                              
Cluster & Description                              & Top 5 terms         \\
9       & First names                              & ? Michael, ?John, ?Emily, ?Robert, ?David                                  \\
17      & Work sectors                             & ?Services, ?Education, ?Technology, ?Engineering, ?Systems                \\
24      & Writing and writing formats              & ?writing, ?email, ?write, ?wrote, ?emails                                 \\
29      & Political names                          & ?Trump, ?Obama, ?Putin, ?Tillerson, ?Clinton                              \\
35      & Artefacts, man-made devices              & ?vehicle, ?laptop, ?device, ?car, ?smartphone                             \\
36      & Generic company names                    & ?Aurora, ?Legacy, ?Alliance, ?Summit, ?Princess                           \\
38      & Official Roles                           & ?Director, ?President, ?Chairman, ?Senator, ?Manager                      \\
39      & Negative and positive terms              & ?frustration, ?sadness, ?anxiety, ?turmoil, ?excitement                   \\
40      & High register terms                      & esteem, terrorism, abortion, commerce, significant                        \\
43      & Colours and flavours                     & ?Blue, ?Chocolate, ?Green, ?Black, ?Beer                                  \\
44      & Tools and construction parts             & ?blade, ?knife, ?rope, ?sewing, ?handle                                   \\
45      & Musical terms                            & ?music, ?songs, ?song, ?musician, ?musicians                              \\
47      & Medical and academic                     & ?Medical, ?Health, ?Researchers, ?Science, ?Environmental                 \\
50      & Municipal                                & ?Mayor, ?municipal, ?mayor, ?council, ?city                               \\
54      & Closed class terms                       & The, And, It, But, That                                                   \\
55      & Past tense negative terms                & ?destroyed, ?arrested, ?attacked, ?defeated, ?detained                    \\
61      & Stages of life                           & ?teenager, ?children, ?kids, ?teenagers, ?babies                          \\
62      & Financial terms                          & ?funding, ?money, ?investments, ?payments, ?revenue                       \\
63      & Religious terms (particularly Christian) & ?church, ?religious, ?churches, ?religion, ?Christians                    \\
72      & Clothing terms                           & ?clothing, ?clothes, ?shirt, ?attire, ?shoes                              \\
73      & Companies                                & ?Microsoft, ?Samsung, ?Google, ?Facebook, ?Netflix                        \\
76      & Medical terms                            & ?medications, ?medication, ?pharmaceutical, ?chemicals, ?bacteria         \\
77      & Sports                                   & ?basketball, ?football, ?tournament, ?soccer, ?baseball                   \\
79      & Measures                                 & ?percent, ?pm, ?mmol, ?tablespoons, \%                                    \\
80      & Group and Organization terms             & ?companies, ?businesses, ?buildings, ?countries, ?areas                   \\
89      & Time zones                               & ?EDT, ?PDT, ?BST, ?GMT, ?EST                                              \\
91      & News                                     & ?newspaper, ?News, ?CNN, ?website, ?newspapers                            \\
92      & Colours and flavours                     & ?orange, ?chocolate, ?wooden, ?purple, ?tomato                            \\
95      & Informal discourse                       & ?yeah, ?Yeah, Yeah, Okay, Oh                                              \\
96      & Ideas                                    & ?notion, ?proposal, ?situation, ?narrative, ?rhetoric                     \\
97      & Buildings                                & ?restaurant, ?hotel, ?apartment, ?museum, ?stadium                        \\
99      & Sex, gender, and sexual terms            & ?sexual, ?women, ?woman, ?feminist, ?female                               \\
100     & Names                                    & John, Michael, Robert, James, Richard                                     \\
101     & Familial relations                       & ?mother, ?father, ?dad, ?parents, ?mom                                    \\
104     & Commerce                                 & ?transformation, ?implementation, ?reduction, ?evaluation, ?inception     \\
106     & Space and science                        & ?spacecraft, ?galaxies, ?solar, ?aerospace, ?physics                      \\
114     & Content                                  & ?videos, ?movies, ?documents, ?films, ?stories                            \\
117     & Years                                    & ?1978, ?1970, ?1977, ?1969, ?1980                                         \\
118     & Geographic terms                         & ?river, ?coastal, ?beach, ?rainfall, ?lake                                \\
119     & Negative events                          & ?Violence, ?Terror, ?Challenge, ?Chaos, ?Fight                            \\
130     & Sports teams                             & ?Seahawks, ?Celtics, ?Falcons, ?Lakers, ?Redskins                         \\
131     & Negative events                          & ?massacre, ?violence, ?confrontation, ?terrorism, ?harassment             \\
135     & Occupations                              & ?journalist, ?lawyer, ?politician, ?reporter, ?businessman                \\
141     & Islam and Islamic countries              & ?Palestin, ?Pakistan, ?Syria, ?Muslim, ?Pakistani                         \\
153     & Claims                                   & ?problems, ?initiatives, ?incidents, ?restrictions, ?discussions          \\
154     & Cities                                   & ?Chicago, ?Philadelphia, ?California, ?Boston, ?Toronto                   \\
156     & Food                                     & ?foods, ?food, ?delicious, ?beer, ?snacks                                 \\
157     & Smell                                    & ?smell, ?smelled, ?smells, ?scent, ?aroma                                 \\
162     & Surnames                                 & ?Gonzalez, ?Lopez, ?Rodriguez, ?Hernandez, ?Cohen                         \\
164     & Maps and directions                      & ?northern, ?southern, ?region, ?country, ?South                           \\
167     & Many                                     & ?both, Both, ?Both, ?between, ?combination                                \\
168     & Functional terms                         & ?regarding, ?adjacent, ?within, ?throughout, ?alongside                   \\
170     & Comparative adjectives                   & ?smaller, ?bigger, ?better, ?stronger, ?larger                            \\
172     & Negative terms                           & ?horrible, ?bizarre, ?ridiculous, ?disgusting, ?terrible                  \\
175     & Positive terms                           & ?incredible, ?fantastic, ?remarkable, ?amazing, ?magnificent              \\
176     & Halloween figures                        & ?vampire, ?monsters, ?vampires, ?superhero, ?dragons                      \\
178     & Military and warfare                     & ?soldiers, ?military, ?firearms, ?weapons, ?troops                        \\
180     & Reports and text                         & Researchers, Reporting, Investigators, Read, Update                       \\
181     & Humour                                   & ?laughed, ?humorous, ?hilarious, ?laugh, ?amusing                         \\
182     & Combination                              & ?partnership, ?collaboration, ?conjunction, ?collaborative, ?collaborated \\
183     & Division                                 & ?split, ?divided, ?separated, ?separating, ?divide                        \\
185     & Countries                                & ?Germany, ?Spain, ?Italy, ?Russia, ?France                                \\
192     & Images                                   & ?photos, ?photo, ?pictures, ?images, ?photographs                         \\
194     & Increases                                & ?increased, ?improve, ?increase, ?enhance, ?improving                     \\
195     & God and deities                          & ?God, God, ?god, ?Jesus, ?deity                                           \\
197     & Body parts                               & ?legs, ?knee, ?neck, ?thigh, ?muscles                                     \\
199     & Animals                                  & ?animals, ?birds, ?cats, ?dogs, ?dog                                     \\
\caption{67 clusters sensitive to semantic information}
\label{tab:sensitive clusters}
\end{longtable}
 \end{landscape} 
 }
\Cref{tab:sensitive clusters} shows how many of the clusters are sensitive to rich information that goes far beyond the surface level features of the expressions. Many of the terms have clustered in a way that is sensitive to their meaning. It is important to note that the distinction between worldly information and semantic information is somewhat blurred in such models. For instance, that a set of names are united by the fact that they pick out political figures (cluster 29) is arguably a fact of the world and not a semantic fact about the names (particularly if we are direct referentialists about names). On the other hand, the fact that e.g. terms are united by their positive sentiment (cluster 175) arguably captures something central about the meaning of such terms. Following Lin \& Murphy’s (2001) distinction between taxonomic and thematic relations, it appears that both types are captured with the static embedding space. For instance, cluster 197 captures a taxonomic relation between terms insofar as the terms fall under the category of bodypart (see also cluster 101 on familial relations). On the other hand, cluster 76 captures medical terms without there being some straightforward hierarchical relation between terms like “medication”, “pharmaceutical”, “chemicals”, and “bacteria”. 

Beyond consideration of the thematic/taxonomic distinction, it is important to note that a wide variety of semantic relations appear to be captured within the space. Clusters like 172 and 175 appear to capture the valence of particular expressions; clusters like 182 and 183 appear to capture some abstract feature shared by the terms (combination and division, respectively); while clusters like 40 and 95 appear to capture the register of particular expressions.

\section{Study 2: Sensitivity to psycholinguistic attributes}\label{sec:study2}

While manual inspection of the clusters allows us to see in detail what the various clusters appear to be sensitive to, we supplement this insight with a quantitative analysis. For our second study, we investigated whether the clusters are sensitive to a range of psycholinguistic measures that either represent or are related to semantic features. Across psycholinguistics, a well-established methodology is to survey participants on the extent to which words possess some feature, and so create a word-list for that feature. While interest within psycholinguistics on such attributes usually concerns their relevance to the processing and cognition of language, we are primarily interested in such measures insofar as they can be taken to stand for or be related to semantic features, even if their ability to do so is admittedly limited. In this study, we focus on valence, concreteness, iconicity, taboo, and age of acquisition. Before detailing our methodology, it will be useful to first describe each attribute.

\subsection{Psycholinguistics attributes}
\subsubsection{Valence}
The \emph{valence} of a term is roughly understood as its pleasantness. The notion is derived from a three-dimensional model of emotional states developed by \parencite{osgoodMeasurementMeaning1957, mehrabianBasicDimensionsGeneral1980}. According to this picture, along with valence, emotions can also vary according to \emph{arousal} (how energetic and attentive an emotion feels) and \emph{dominance} (how active or passive the emotion feels). We have focused only on valence, as it is often taken as the most significant dimension of the three \parencite[1192]{warrinerNormsValenceArousal2013} and is arguably also the most intuitive. Here we use Warriner et al.'s \parencite*{warrinerNormsValenceArousal2013} word list for 13,915 English lemmas.

\subsubsection{Concreteness}
Concreteness stands for the extent to which a word refers to a perceptible entity rather than an abstract notion. So “bicycle” would have a high concreteness score (4.89) while “justice” would have a low concreteness score (1.45). Here we use Brysbaert et al.'s \parencite*{brysbaertConcretenessRatings402014} scores for 37,058 English words. 

\subsubsection{Iconicity}
A view once widely-held is that the relationship between a word’s iconographic and phonological features on the one hand, and its semantic features on the other, is arbitrary, bar a few exceptions of onomatopoeia (e.g. “boom”, “fizzle”). More recently, this position has been challenged by the idea that iconicity - a “perceived resemblance between aspects of [..] form and aspects of [...] meaning” \parencite[1640]{winterIconicityRatings140002024} actually appears across a wide array of terms to varying extents. Iconicity is an intriguing property to consider with regard to LLMs because to detect iconicity, three things are obviously needed. First, you need access to a word’s surface properties (phonological, iconographic, etc.), second you need access to a word’s semantic properties, and third you need to recognize a resemblance between the two. It is important to note that neither the surface properties nor the semantic properties are explicitly made available to an LLM, these are features that would have to be inferred through training. So it would be a further feat still if the LLM has not only encoded these properties within the representation of such words, but also encoded that there is a resemblance between them. Here we use Winter et al.’s \parencite*{winterIconicityRatings140002024} list for 14,776 English words.

\subsubsection{Taboo}
A type of meaning that has been of interest to philosophers of language in recent years is pejorative and slur meaning. As a form of meaning, it appears to behave uniquely insofar as the offensive content is still communicated even when such expressions are embedded in conditional sentences, speech act reports, and other sentential contexts. As this kind of meaning appears to invariantly be communicated by the expressions, this kind of meaning seems like a good candidate for the kind of information that would be encoded within static embeddings. Here we draw upon Reilly et al.’s \parencite*{reillyBuildingPerfectCurse2020} taboo list for 1,205 words. Tabooness is not equivalent to the category of either slurs or pejoratives. Words like “dildo” and “goddam” have relatively high taboo scores but are not clearly pejoratives and certainly not slurs. Following Reilly et al., we take tabooness to track something like the extent to which a word is a “swear” word or “curse” word. 

\subsubsection{Age of Acquisition}

Age of acquisition (AoA) is the age at which a word is acquired. This is a widely-studied attribute in psycholinguistics due to the fact that AoA has been shown to correlate strongly with processing time. While not obviously a semantic feature itself, AoA is nevertheless an interesting attribute to test for in static embeddings, as it has been shown that AoA correlates to some extent with certain semantic features, such as imagability - the ease with which a word gives rise to a mental image \parencite{birdAgeAcquisitionImageability2001}. As age of acquisition also correlates with frequency, we do not think that showing that the static embedding clusters are sensitive to AoA would on its own establish that semantic information is encoded. But when grouped with the four other attributes, we take a sensitivity to AoA to further strengthen the case. Here we use Kuperman et al.'s \parencite*{kupermanAgeofacquisitionRatings300002012} list for 30,121 words. 

\subsection{Method}
Before testing for sensitivity to each attribute, we first ensured that duplicate values were not generated. Because words are identified typographically in the tokenisation procedure, and because a word is assigned a distinct embedding depending on whether it appeared at the start of a sentence or halfway through, there are many duplicates within RoBERTa’s vocabulary. For example, the word “dog” will have many entries: “Dog”, “ Dog”, “DOG”, “dog”, “ dog”  etc. In assigning attribute scores to words in each cluster, we decided to assign a value to only one of these duplicated entries. A case could be made for not doing so: prior to any training the model treats the above typographical variations as completely distinct tokens with distinct embeddings and so it is worth seeing what kind of information has been stored in each. However, to avoid any charge of inflating our results, we took the more conservative approach. For each attribute value, we assign it either to the first case-matched entry, or where there is no case-matched entry, to the first non-case-matched entry. As a result, the number of entries with attributes assigned are given in \cref{tab:assigned tokens}.

\begin{table}[]
\caption{Tokens assigned values for each attribute}
\label{tab:assigned tokens}
\begin{tabular}{|>{\centering\arraybackslash}p{2cm}|>{\centering\arraybackslash}p{2cm}|>{\centering\arraybackslash}p{2cm}|>{\centering\arraybackslash}p{2.5cm}|>{\centering\arraybackslash}p{2.5cm}|}
\hline
Attribute & Word-list length & Tokens assigned & Case-sensitive matches & Case-insensitive matches \\
\hline
Valence & 13915 & 8751 & 6977 & 1774 \\
\hline
Concreteness & 37058 & 12334 & 9833 & 2501 \\
\hline
Iconicity & 14776 & 8909 & 6933 & 1976 \\
\hline
Taboo & 1205 & 1085 & 848 & 237 \\
\hline
AoA & 30121 & 10574 & 8299 & 2275 \\
\hline
\end{tabular}
\end{table}

For each attribute, we test whether each cluster is organized in a way that is sensitive to that attribute. More specifically, we investigate the probability of the distribution of that attribute across the cluster given the distribution of that attribute across the entire dataset.  

We proceed as follows. For each attribute, we group the range of values according to their integral parts. For instance, if we consider the valence attribute, we end up with integral part values 1-8. Counts of the numbers of valence annotated words within each integral part value are given by:
$${\bf c} = \{53, 576, 984, 1740, 3021, 1760, 587, 30\}.$$
We let: $$p_{\rm cat}(j)=\{0.00606, 0.0658, 0.112, 0.199, 0.345, 0.201, 0.0671, 0.00343\}$$ denote the probability that a randomly drawn  word (from those with a valence attribution) lies  within  valence category  $j$, for $j=1,...,8$. These probabilities sum to unity, of course, being derived directly from the categorical counts given in ${\bf c}$.  Similarly,  we let $p_{\rm clust}(i)$ denote the  probability that a randomly drawn word (with a valence attribution) lies within cluster $i$, for $i=1,...,200$.

Now  we let $p(i,j)$ denote the probability that a randomly drawn  word (with a valence attribution)  lies within   cluster $i$ and  valence category $j$,  $i=1,...,200$ and $j=1,..,8$. The $p_{\rm clust}$ and $p_{\rm cat}$  probability distributions  are called  the marginal (categorical) probability distributions, while the $p(i,j)$ denote the joint (categorical) probability distribution. Then we can calculate  the entropies,  of each categorical distributions and the (normalised) mutual information of the two categorical distributions.

In this case we find the (normalised) mutual information is low, indicating that for an average word drawn from the total population, distributed over the $200\times 8$ grid,  there is little difference between the joint probability and the corresponding  product of the marginals. This is because within most of the clusters there is no great effect on the way that the valence categories are distributed (those clusters are valence independent). However, in fact some of the clusters are indeed related to highly skewed distributions of valence, and there the joint distribution is highly distinct from the product of the marginals.

Consider cluster 172 containing $m=340$ words, for example. The counts of the number of words for each valence categorical value are given by:
$$C =\{4, 117, 149, 52, 12, 6, 0, 0\}$$ 
which appear to be distributed very differently from the total population counts, ${\bf c}$,  given above (which formed the basis of the marginal $p_{\rm cat}$). 

In fact, we can measure this by calculating the (natural) log probability of observing $C$, given the marginal distribution $p_{\rm cat}$, when drawing $m=340$ words.  That is, by assuming the Null Hypothesis (NH) that the words within cluster 172 are merely a random set drawn from whole population, with a distribution given by the marginal, $p_{\rm cat}$. The conditional probability of observing the actual cluster 172  counts, $C$,  under the NH is denoted by  $P(C|p_{\rm cat})$, where:
$$P(C|p_{\rm cat})= \frac{m!}{C_1! C_2!...,C_{8}!}\Pi_{i=1}^{8} p_{\rm cat}(i)^{C_i}
.$$ 
Then (taking natural logarithms so as to deal with extremely low probabilities), in this case, we have:
$$\log P(C|p_{\rm cat})=\sum_{i=1}^{8} C_i \log p_{\rm cat}(i) + \log m! -\sum_{i=1}^{8} \log C_i! = -354.667.$$

\begin{figure}[]
\centering
\includegraphics[width=0.85\textwidth]{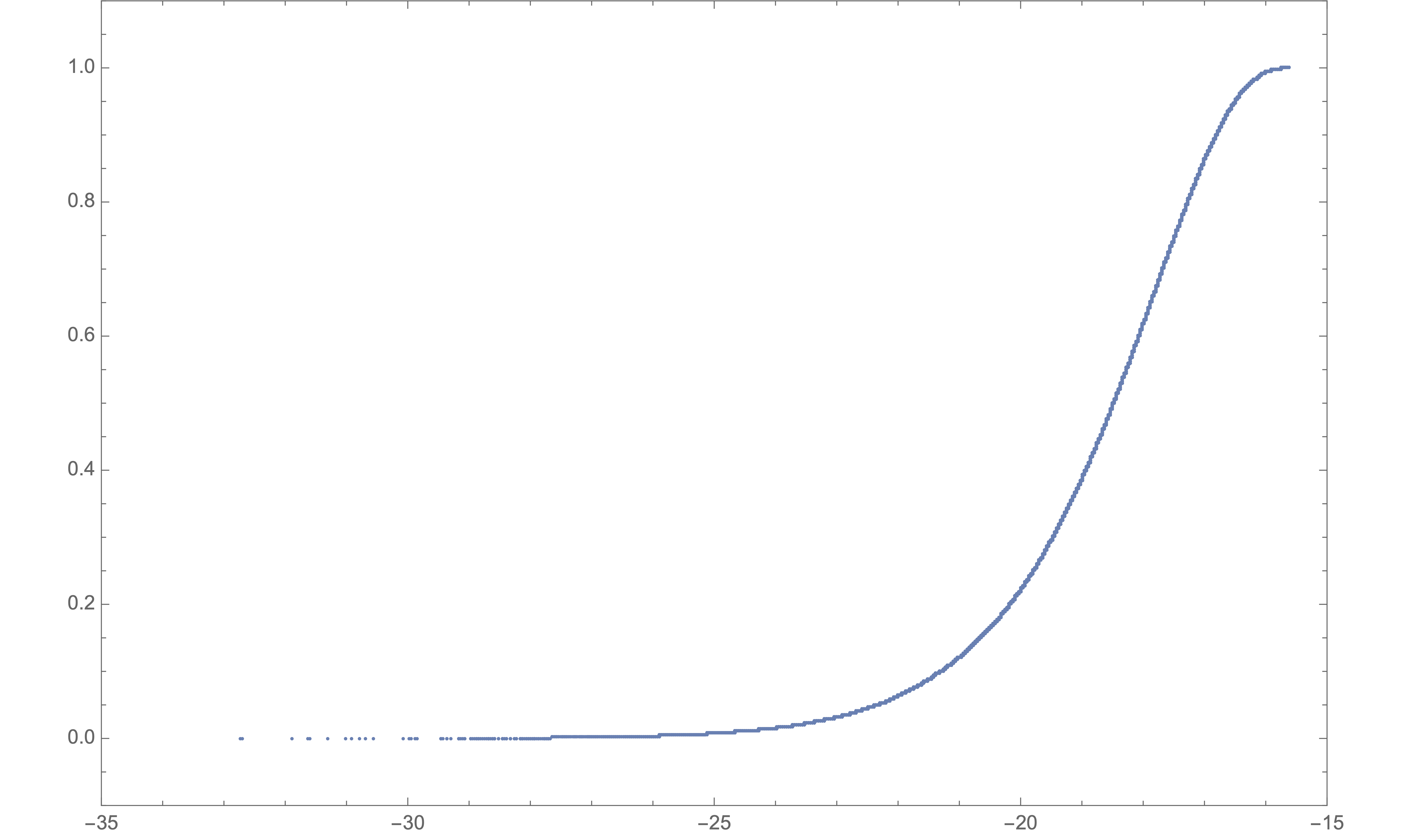}
\caption{Cummulative distribution  of $\log P(\tilde{C}|p_{\rm cat})$ values achieved conditional on  the valence   attribute distribution (that is conditional on assuming the NH)  from 100,0000 independent  samples, each of  the same size as  cluster 172 ($m=340$). In fact the corresponding  $\log P$ value achieved by the actual cluster 172 subset of words is -354.667, which is extremely low. }
\label{fig:valenceCUM3}
\end{figure}

On the other hand, if we actually select random sets of the same size, $m=340$, according to the marginal valence distribution, $p_{\rm cat}$,  resulting in different counts, say $\tilde{C}$, we may calculate  the analogous statistic $\log P(\tilde{C}|p_{\rm cat})$ values. Over 100,000 such independently sampled subsets  we show the range of the $\log P(\tilde{C}|p_{\rm cat})$  values as a cumulative distribution   in Figure \ref{fig:valenceCUM3}. The actual value achieved by the Cluster 172 subset of words is -354.667, which lies very far beneath the ``range'' of those values achievable assuming the NH: we will set the attainable range to be such that the probability, $p$,  of any set of size $m$ drawn under the NH sitting below it satisfies  $p << 5/100000$.

\subsection{Results}

\begin{figure}[]
\centering
\includegraphics[width=0.48\textwidth]{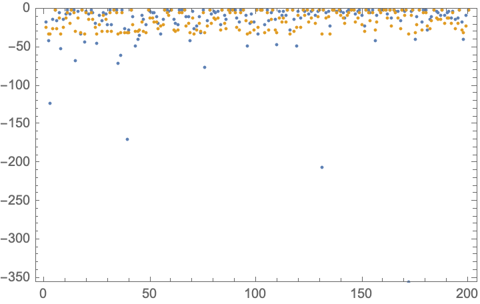}
\hspace{0.1cm}
\includegraphics[width=0.48\textwidth]{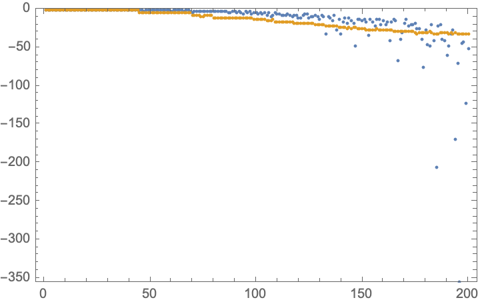}
\caption{Results for the valence attribute (clusters ordered by size on the right-hand side)}
\label{fig:valence}
\end{figure}

The values for $\log  P({C}|p_{\rm cat})$ are shown for all 200 clusters and valence in \cref{fig:valence} (in blue). For each cluster we also calculated the equivalent distribution of value for 100,000 similar sized random subsets drawn under the NH (in yellow).\footnote{ We are using natural logarithms in these calculations.} In \cref{fig:valence}, any blue dot that falls below its yellow dot is lower than $p=0.00005$, and so the NH is disconfirmed for that cluster and attribute.

\begin{figure}[]
\centering
\includegraphics[width=0.48\textwidth]{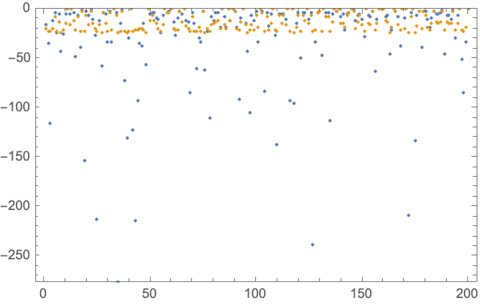}
\hspace{0.1cm}
\includegraphics[width=0.48\textwidth]{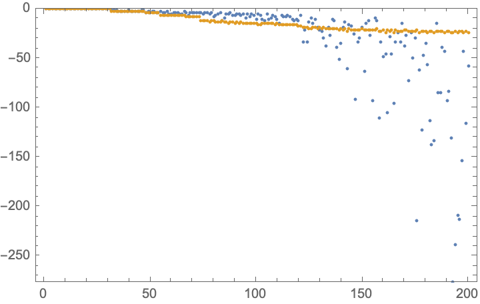}
\caption{Results for concreteness}
\label{fig:Concrete}
\end{figure}

Next  consider the concreteness attribute, where we have $12,334$ annotated tokens, and we partition the concreteness values into 5 bins, taking the integer parts of the values in [1.04, 5].  This had many more sensitive clusters, 60 in total. The results are shown in \cref{fig:Concrete}. 

\begin{figure}[]
\centering
\includegraphics[width=0.48\textwidth]{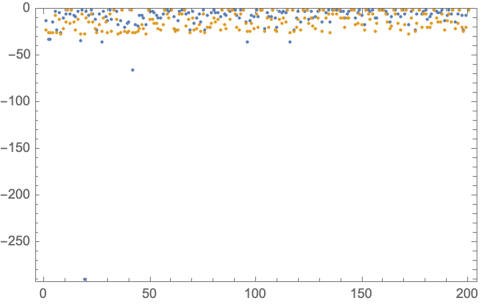}
\hspace{0.1cm}
\includegraphics[width=0.48\textwidth]{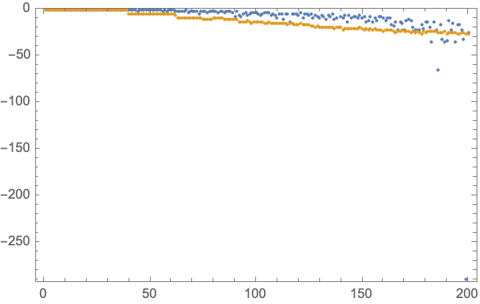}
\caption{Results for iconicity}
\label{fig:iconicity}
\end{figure}

For iconicity, we have $8,909$ annotated tokens, which we partition into 6 bins, by taking the integer parts of values between [1.3, 6.727272727272728]. Although this is a similar number of annotations to valence, there were far fewer sensitive clusters, with only 9 in total. Results are given in \cref{fig:iconicity}.

\begin{figure}[]
\centering
\includegraphics[width=0.48\textwidth]{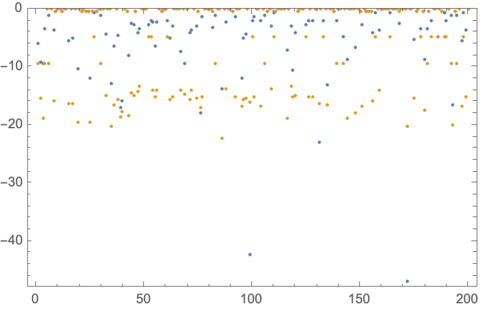}
\hspace{0.1cm}
\includegraphics[width=0.48\textwidth]{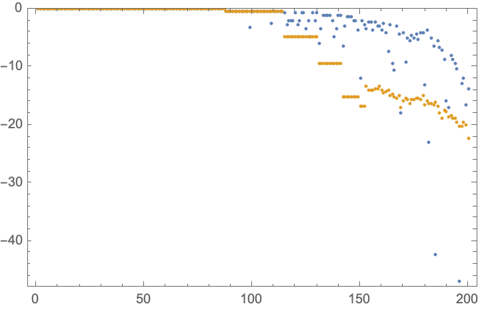}
\caption{Results for taboo}
\label{fig:taboo}
\end{figure}

For Taboo, we had relatively few annotations ($1,084$) which were partitioned into 7 bins, ranging from [1, 7.333333333]. The null hypothesis was disconfirmed for 6 clusters, but for 2 of these clusters this was really just the result of an extreme sampling error.  Clusters 82 and 168 only had one token each annotated with a taboo value, so these results can be discounted. Results are given in \cref{fig:taboo}. 

\begin{figure}[]
\centering
\includegraphics[width=0.48\textwidth]{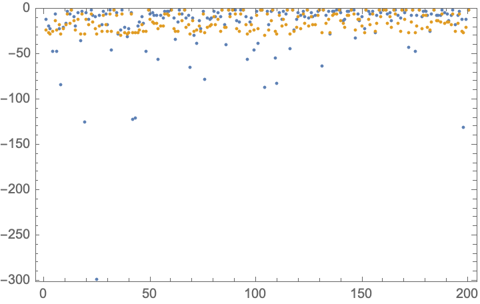}
\hspace{0.1cm}
\includegraphics[width=0.48\textwidth]{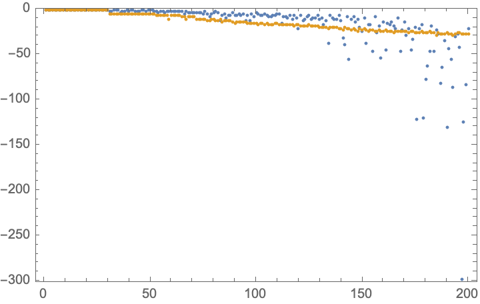}
\caption{Results for AoA}
\label{fig:Aoa}
\end{figure}

Finally, AoA had $10,574$ tokens assigned, which we partitioned into 18 bins ranging from [1.89, 18.52]. 36 clusters were sensitive to AoA, with results given in \cref{fig:Aoa}. One of the clusters, cluster 149, had only 5 AoA values, and the cluster appears to only consist in sub-word parts, so this cluster should be discounted. 

All 6 attributes had some clusters that were sensitive to them, although they differed markedly in terms of the number of clusters (\cref{tab:clustersoverview}). Notably, iconicity and taboo had far fewer sensitive clusters than the other attributes. 73 clusters are sensitive to at least one attribute. Some clusters are sensitive to a number of attributes, with 6 clusters sensitive to 4 attributes, although none are sensitive to all (\cref{tab:attributenumbers}).

\begin{table}[h]
\centering
\caption{Number of clusters sensitive to each attribute}
\label{tab:clustersoverview}
\begin{tabular}{ll}
Attribute    & Clusters \\
Valence      & 27       \\
Concreteness & 60       \\
Iconicity    & 9        \\
Taboo        & 6        \\
AoA          & 36      
\end{tabular}
\end{table}

\begin{table}[h]
\centering
\caption{Distribution of sensitive clusters across number of attributes}
\label{tab:attributenumbers}
\begin{tabular}{ll}
Attributes & Count \\
5          & 0     \\
4          & 6     \\
3          & 15    \\
2          & 17    \\
1          & 35    \\
0          & 127  
\end{tabular}
\end{table}

\subsection{Discussion}

The results from study 2 add further reason to think that the static embeddings encode a wide array of semantic information. It is notable, for instance, that we even found clusters to be sensitive to taboo, even though the tokens assigned a taboo score was relatively small, and there is \emph{a priori} reason to think that taboo is a relatively marginal or niche form of meaning, contained to a relatively small set of expressions. At the other end of the scale, attributes like concreteness, and to a lesser extent valence and AoA, appear to have had an effect on the organisation of a relatively large number of clusters. 

There is some reason to be cautious, however. For any given cluster that has an improbable distribution of a given attribute, there is still the possibility that the clusters in question are not actually sensitive to the attribute, but are sensitive to some correlative property. It is certainly not the case that a sensitive cluster should be treated as about that attribute (or the absence of that attribute). For instance, clusters 76, 99, and 131 are sensitive to taboo, but were labelled in study 1 as about ``medical terms'', ``sex, gender, and sexual terms'', and ``negative events'' respectively.  While we do not take this study to identify clusters that are best-described as united by the attributes we tested for, the fact that the clusters are sensitive to such attributes nevertheless still supports the claim that a range of semantic information is encoded at the static embedding level, as it is not plausible to think that clusters organized only by, say, syntactic or morphological features could correlate attributes of this kind.

We are a little more skeptical, however, of the iconicity findings. Only 9 clusters were sensitive to iconicity, despite a large number of annotated tokens, and manual inspection of the sensitive clusters does not reveal any that clearly capture iconicity or some related feature. There is the possibility that iconicity correlates with some surface feature such as number of syllables (with one and two syllable words being more likely to have high iconicity) and so some of the clusters in question are sensitive to that. This is not something we will explore further here, but merely register it as a limitation of the study. 

We also found that running such a test with an attribute list as small as taboo (only $1,085$), led to sampling errors for certain clusters (notably clusters 68 and 82), and likely contributed as well to the small number of clusters that tested as sensitive to taboo. However, we don't take this to discredit the taboo findings entirely. The other clusters that tested as sensitive to taboo do, upon manual inspection, admit of plausible explanations as to why they would be sensitive. For instance, cluster 76 appears to be about medical terms, but as a result has terms like ``cancer'', ``tumour'', and ``heroine'', all of which plausibly have a high taboo value. 

The pipeline methodology we have employed involved word embedding within a high dimensional Euclidean space based on pairwise proximities, unsupervised clustering, and statistical tests for psycholinguistic attribute distributions within clusters. The methodology could be generalised in many ways. For example,  use of alternative embeddings; use of variations of the  EM algorithm, rather than simple K-means clustering; use of alternative null hypotheses. It may be worth considering subsets of the corpora partitioned  by {\it source }classifications. For example, cultural  biases might support  alternative uses of language by distinct sub-population groups (based on age, culture, ethnicity) or within distinct settings (legal, formal, media, casual, genre). It could be important to test this as Simpson's paradox is always lurking,  and possibly masking idiosyncratic meanings within well-defined, yet narrower settings and sub-populations. But despite these possible avenues of future research, we take our results here to point in a clear direction. They confirm the claim that semantic information is encoded at the static embedding level. This serves to reject the meaning eliminativist picture of LLMs that we outlined in \cref{sec:Ceac}. According to that picture, the LLM has no need to invariantly associate with individual expressions semantic information, as this is something that can be realized once the model has taken the sentential context into account. In contrast to that picture, we have found that LLMs do encode semantic information at the static embedding level, even though these embeddings subsequently go through a massive set of transformations with no \emph{a priori} constraints set on them. LLMs do still require a lexical store of semantic information as part of their procedure for understanding text. 

\section{Conclusion}
We have shown in this paper that the static embeddings that serve as input to the self-attention procedure do not merely store syntactic and surface-level information about words (and word parts) but also store meaningful semantic information. We have also seen some reason to think that worldly information is stored at this level as well (for instance, cluster 29 captures political names while cluster 130 captures North American sports team names). In general, we have sought to employ a relatively simple methodology in order to consider empirical questions about how transformer models operate in a way that can speak to debates in philosophy of language and linguistics regarding the nature of meaning, understanding, and communication.

\section*{Declarations}
PG's research  was funded by UKRI EPSRC grant number EP/Y007484/1 {\it  Mathematical Foundations of Intelligence}. \\ 
For the purpose of open access, the authors have applied a CC BY public copyright licence to any Author Accepted Manuscript version arising from this submission.

\printbibliography
\end{document}